\documentclass{article}

\usepackage{microtype}
\usepackage{graphicx}
\usepackage{subfigure}
\usepackage{booktabs} 

\usepackage{hyperref}

\usepackage[accepted]{icml2020}

\usepackage{amsfonts}
\usepackage{tabularx}
\usepackage{amsmath}
\usepackage{amsfonts}
\usepackage{tabularx}
\usepackage{caption}
\usepackage[flushleft]{threeparttable}
\newcolumntype{R}{>{\raggedleft\arraybackslash}X}

\icmltitlerunning{Regression via Implicit Models and Optimal Transport Cost Minimization}

\begin{document}

\twocolumn[
\icmltitle{Regression via Implicit Models and Optimal Transport Cost Minimization}




\begin{icmlauthorlist}
\icmlauthor{Saurav Manchanda}{to}
\icmlauthor{Khoa Doan}{goo}
\icmlauthor{Pranjul Yadav}{ed}
\icmlauthor{S. Sathiya Keerthi}{edk}
\end{icmlauthorlist}

\icmlaffiliation{to}{Department of Computer Science and Engineering, University of Minnesota, Twin-Cities, MN, USA}
\icmlaffiliation{goo}{Department of Computer Science, Virginia Tech, Arlington, VA, USA}
\icmlaffiliation{ed}{Criteo AI Lab, Palo Alto, CA, USA}
\icmlaffiliation{edk}{LinkedIn Corporation, Mountain View, CA, USA}

\icmlcorrespondingauthor{Saurav Manchanda}{manch043@umn.edu}

\icmlkeywords{GAN, WGAN, Wasserstein Gan, Adversarial regression}

\vskip 0.3in
]



\printAffiliationsAndNotice{}  
\begin{abstract}
This paper addresses the classic problem of regression, which involves the inductive learning of a map, $y=f(x,z)$, $z$ denoting noise, $f:\mathbb{R}^n\times \mathbb{R}^k \rightarrow \mathbb{R}^m$. Recently, Conditional GAN (CGAN) has been applied for regression and has shown to be advantageous over the other standard approaches like Gaussian Process Regression, given its ability to implicitly model complex noise forms. However, the current CGAN implementation for regression uses the classical generator-discriminator architecture with the minimax optimization approach, which is notorious for being difficult to train due to issues like training instability or failure to converge. In this paper, we take another step towards regression models that implicitly model the noise, and propose a solution which directly optimizes the optimal transport cost between the true probability distribution $p(y|x)$ and the estimated distribution $\hat{p}(y|x)$ and does not suffer from the issues associated with the minimax approach. On a variety of synthetic and real-world datasets, our proposed solution achieves state-of-the-art results\footnote{The code accompanying this paper is available at \url{https://github.com/gurdaspuriya/ot_regression}}.

\end{abstract}

\section{Introduction}
\label{sec:intro}
Regression analysis estimates the relationships between one or more dependent variables (the `outcome variables', denoted by $y\in\mathbb{R}^m$) and one or more independent variables (the `features', denoted by $x\in\mathbb{R}^n$). As such, it is one of the fundamental problems in machine learning and statistics. Formally, regression involves estimating $y$ as a continuous and stochastic function of $x$, i.e., regression is the inductive learning of a map $y=f(x,z)$, $z\in\mathbb{R}^k$ denoting noise.

A common assumption made by most of the regression approaches~\cite{bishop2006pattern} is to model the noise in an additive fashion, i.e.,
\begin{equation}
y = \hat{f}(x) + z
\end{equation}
where $z$ is a parametric noise. For example, (i) Regression using Gaussian processes (GPs)~\cite{williams1996gaussian} assumes a zero-mean Gaussian noise, i.e., $z \sim \mathcal{N}(0,\Sigma)$, where $\Sigma$ is the covariance matrix; and (ii) Heteroscedastic Gaussian Process Regression (GPs)~\cite{le2005heteroscedastic} extends the Gaussian Process Regression and assumes a zero-mean Gaussian noise with the covariance matrix dependent on $x$, i.e., $z|x \sim \mathcal{N}(0,\Sigma |x)$, where $\Sigma |x$ is the \emph{local} covariance matrix. Such assumptions are made so as to have the posterior $p(y|x)$ in a closed form.

Such assumptions regarding how the noise enters the true $y$-generation may not hold in real world systems. For example, in such systems, the noise can arise from several sources, such as failure of measurement devices, leading to missing values; and other uncontrolled interferences. As such, without having the apriori information about the noise sources, it is unreasonable to assume how the noise enters the true $y$-generation process. In this paper, we present non-parametric approaches that implicitly model the noise, without any assumption about the noise. 

Recently, Conditional GAN (CGAN)~\cite{aggarwal2019regression} has been employed for regression, which is also a non-parametric approach that implicitly models the noise. However, the current CGAN implementation for regression employs the original minimax GAN objective~\cite{goodfellow2014generative} to discriminate the samples drawn from the true probability distribution $p(y|x)$ and samples drawn from the estimated distribution $\hat{p}(y|x)$. This minimax objective suffers from various challenges, such as mode-collapse and vanishing gradient~\cite{arjovsky2017wasserstein}. Moreover, in minimax optimization, the generator's loss fluctuates during the training instead of ``descending'', making it extremely challenging to know when to stop the training process. In contrast, our proposed solution models the noise implicitly, but without having the limitations of the minimax optimization. Specifically, we formulate our problem as a minimization of the optimal transport cost between the true probability distribution $p(y|x)$ and the estimated distribution $\hat{p}(y|x)$. This formulation boils down to a solution of a linear-assignment problem and does not suffer from the issues associated with the minimax approach. 

The optimal transport cost is also related to the \emph{Wasserstein} distance between the two distributions (as explained in Section~\ref{sec:background}). GANs that use the Wasserstein distance (Wasserstein GANs or WGANs) have worked well in image applications. The usual approach in WGAN is to use duality and the associated minimax solution with controlled steps. Thus, the usual approach for WGAN still faces the issues that occur as a result of the minimax optimization. The key reason for using the dual approach in image applications is not just computational ease: the primal WGAN that is based on using Euclidean/Laplacian distances in the original pixel feature space is simply unsuitable for images; by using suitable image processing friendly architectures to express the dual function, the problem is avoided. In regression there are no such issues; the output space, i.e., the cardinality of $y$ is usually small, and Euclidean/Laplacian distances in the original space is suitable. This allows us to go for a direct, primal Wasserstein distance optimization, which takes the form of a simple linear programming objective. This is a key element of our work that helps avoid all the mess related to the minimax approach.

We choose several synthetic and real world datasets from the various types of regression problem domains for evaluation. Experiments demonstrate that our proposed approach achieves state-of-the-art performance on all the datasets. 

\section{Background}\label{sec:background}

\begin{table}[!t]
\small
\centering
  \caption{Notation used throughout the paper.}
  \begin{tabularx}{\columnwidth}{lX}
    \hline
Symbol   & Description \\ \hline
$x$    & Input features, $x \in \mathbb{R}^n$. \\
$y$    & True output, $y \in \mathbb{R}^m$. \\
$z$    & Noise, $z \in \mathbb{R}^k$. \\
$f$    & Regression function to be estimate. \\
$\theta$    & Parameters of the function $f$. \\
$\mathbb{P}_R$    & Actual joint distribution of $x$ and $y$, i.e, $p(x,y)$. \\
$\mathbb{P}_F$    & Predicted joint distribution of $x$ and $y$, i.e, $\hat{p}(x,y)$.\\
$R$    & Set of samples drawn from the distribution $\mathbb{P}_R$. \\
$F$    & Set of samples drawn from the distribution $\mathbb{P}_F$. \\
$N$    & Number of samples in $R$ (and $F$). \\
$c(a,b)$    & Unit transport cost from $a$ to $b$. \\
$C(\mathbb{P}_R, \mathbb{P}_F)$ & Optimal transport cost between the $\mathbb{P}_R$ and $\mathbb{P}_F$. \\
$\hat{C}(R, F)$ & Empirical transport cost between the samples $R \sim \mathbb{P}_R$ and $F \sim \mathbb{P}_F$. \\
$M$ & Transport plan, where $M_{a,b} = 1$ is corresponding to transporting mass from $a\in R$ to $b\in F$. \\
$M^*$ & Optimal transport plan. \\
$\lambda$ & Hyperparameter controlling the contribution of $x$ and $y$ towards the cost function, for the optimal transport problem. \\
\hline
\end{tabularx}
  \label{tab:notation}
\end{table}

The core of the problem that we address in the paper is to estimate the distance (or closeness) between two distributions $\mathbb{P}_R$ and $\mathbb{P}_F$ using only the samples drawn from these distributions.\footnote{Our use of $R$ and $F$ comes from GAN terminology: $R$ corresponds to the real (true) distribution, while $F$ corresponds to the fake (predicted) distribution.} We estimate this distance as the optimal transport cost $(C(\mathbb{P}_R, \mathbb{P}_F))$ between $\mathbb{P}_R$ and $\mathbb{P}_F$ as follows:
\begin{equation}
    C(\mathbb{P}_R, \mathbb{P}_F) = \inf_{\gamma \in \pi(\mathcal{P}_R, \mathcal{P}_F)}  \mathbb{E}_{(a,b)\in\gamma}[c(a,b)],
\end{equation}
where $c(a, b)$ is the cost of transporting one unit mass from $a$ to $b$ and is assumed to be differentiable with respect to its arguments almost everywhere, and $\pi(\mathbb{P}_R, \mathbb{P}_F)$ denotes the set of all couplings between $\mathbb{P}_R$ and $\mathbb{P}_F$, that is, all joint probability distributions that have marginals $\mathbb{P}_R$ and $\mathbb{P}_F$.

As outlined in~\cite{iohara2018generative}, the optimal transport cost between two probability distributions, $\mathbb{P}_R$ and $\mathbb{P}_F$ can be evaluated efficiently when the distributions are uniform over finite sets of the same cardinality. Specifically, let $R = \{a: a\sim\mathbb{P}_R\}$ denote a set of samples drawn from the probability distribution $\mathbb{P}_R$. Similarly, let $F = \{b: b\sim\mathbb{P}_F\}$ denote the set of samples drawn from the probability distribution $\mathbb{P}_F$, such that $|R| = |F| = N$. By a slight abuse of notation, we will use $R$ and $F$ to denote the finite sample sets as well as the corresponding empirical distributions. The empirical optimal transport cost between the distributions $R$ and $F$ is given by:
\begin{equation}\label{eq:lp}
\begin{matrix}
\hat{C}(R, F)& = &\frac{1}{N}\displaystyle \min_M\displaystyle \sum_{a\in R}\displaystyle \sum_{b \in F} M_{a,b} c(a,b)  \\
&\textrm{s.t.} & \displaystyle \sum_{b\in F} M_{a,b} = 1 \quad\forall a\in R  \\
& & \displaystyle \sum_{a\in R} M_{a,b} = 1 \quad\forall b\in F  \\
& &  M_{a,b} \geq 0 \quad\forall a\in R; \quad  b\in F  .
\end{matrix}
\end{equation}
Further, the solution $M^*$ to the above linear-programming problem is a permutation matrix. Thus, the constraints in (\ref{eq:lp}) automatically imply that $M^*_{a,b} \in \{0,1\} \; \forall (a,b)$ where $M^*$ is the optimal transport plan, and $M^*_{a,b} = 1$ corresponds to transporting the mass from $a\in R$ to $b\in F$. The full set of notations used in this paper is presented in Table~\ref{tab:notation}. 

\textbf{Relationship to Wasserstein-GAN (WGAN)}: The Wasserstein-p distance $(W_p(\mathbb{P}_R, \mathbb{P}_F))$ between the probability distributions $\mathbb{P}_R$ and $\mathbb{P}_F$ can be defined in terms of the optimal transport cost $(C(\mathbb{P}_R, \mathbb{P}_F))$ between $\mathbb{P}_R$ and $\mathbb{P}_F$ with $c(a,b) = ||a-b||^p$ as 
\begin{equation}
    W_p(\mathbb{P}_R, \mathbb{P}_F) = C(\mathbb{P}_R, \mathbb{P}_F)^{1/p}.
\end{equation}
Wasserstein GAN (WGAN) has worked well in image applications. Due the high intractability of the primal as well as other reasons mentioned in section~\ref{sec:intro}, WGAN employs the Kantorovich duality and the associated minimax solution, with controlled steps on the generator and discriminator. Thus it still faces the issues that occur as a result of the minimax optimization. 
The minimax optimization procedure is also slow and possibly requires hand-holding for different applications. However, for regression, as we  show through our experiments, Euclidean/Laplacian distances in the original space is suitable, and hence, this allows us to go for a direct Wasserstein distance optimization of the linear programming objective in (\ref{eq:lp}).

\section{The Model}\label{sec:model}
The task at hand is regression, where the true output ($y\in \mathbb{R}^m$) is a continuous and stochastic function of the input ($x\in \mathbb{R}^n$): $
y = f(x,z) \;\; \mbox{where} \;\; z\in \mathbb{R}^k \;\; \mbox{is the noise vector.} 
$
We take $z$ to come from a known distribution, e.g., the standard Normal.
Specifically, we want to estimate this function $f$, that takes $x$ and $z$ as input, and generates the output $y$. For the same $x$, different values of $y$ can be generated corresponding to different noise $z$; thus allowing us to draw samples from $p(y|x)$. As discussed in Section~\ref{sec:intro}, most of the prior approaches for regression make additional assumptions about the distribution $p(y|x)$. For example, mean squared error (MSE), the popular regression loss, assumes that $p(y|x)$ is normally distributed. Specifically, MSE comes from the cross entropy between the empirical distribution and a Gaussian distribution~\cite{goodfellow2016deep}. We intend to look into models that can implicitly model the noise $z$ without any assumption on how the noise enters the $y$-generation process. 

We denote the joint probability distribution $p(x,y)$ of the training data as $\mathbb{P}_R$. The joint probability with respect to the predicted $y$ is denoted by $\mathbb{P}_F$. 
To model $\mathbb{P}_R$, we use the $N$ samples of $\{x, y\}$ in the training data, and denote the set of these samples by $R$. Also, we take the $x$ values in the training data to represent $p(x)$ and use the function $f$ and the known distribution $z$ to generate $N$ samples of $\{x, y\}$ to model $\mathbb{P}_F$, and denote the set of these samples by $F$. 

Note that, to estimate the function $f$, we want to model the conditional distribution $p(y|x)$, and not the joint distribution $p(x,y)$. However, since $x$ and $y$ are continuous, it is difficult for the real world datasets to have the same $x$ repeated; thus we expect the real world datasets to contain only a single sample drawn from the real conditional distribution $p(y|x)$ for each $x$. This makes directly estimating $p(y|x)$ a non-trivial task. However, as $y$ is a continuous function of $x$, $p(y|x)$ should not change much in the neighborhood of $x$. Thus, by minimizing the optimal transport cost between the $R$ and $F$, but by constraining that the transport plan maps $a \in R$ to $b \in F$, only if $a$ and $b$ lies in the $x$-neighborhood of each other, we implicitly supervise $f$ to model $p(y|x)$. 

As discussed earlier, this optimal transport cost is formulated as a linear assignment problem.
Given the optimal transport cost $\hat{C}^*(R, F)$, we need its gradient\footnote{$\hat{C}^*(R, F)$ is actually a piecewise differentiable function of $\theta$. Under weak assumptions one can show that the gradient exists for almost all $\theta$.} with respect to the parameter vector $\theta$ of the function $f$, to update it. The optimal transport plan $M^*$ is unchanged under small enough perturbations of $\theta$; using this, it is easy to show that the gradient of the optimal transport cost with respect to $\theta$ is
\begin{equation}
    \partial_\theta C = \frac{1}{N}\displaystyle \sum_{a\in R}\displaystyle \sum_{b \in F} M^*_{a,b} \times \partial_b c(a,b) \times \partial_\theta b.
\end{equation}
As mentioned before, the transport plan should should only map $a$ to $b$, if $a$ and $b$ lie close in the $x$-space. Thus, the requirements for the cost function $c(a,b)$ are:
\begin{itemize}
    \item $c(a,b)$ should be relatively high when $b$ does not lie in the $x$-neighborhood of $a$;
    \item $c(a,b)$ should be piecewise differentiable with respect to $\theta$.
\end{itemize}
In this paper, we model $c(a,b)$ using the weighted $L_p$ distance as:
\begin{multline}\label{eq:cost}
    c(a,b) = c(\{x_a,y_a\}, \{x_b,y_b\}) \\= [\frac{\lambda}{n}\sum_{i=1}^n|x_{a,i} - x_{b,i}|^p + \frac{1-\lambda}{m}\sum_{i=1}^m|y_{a,i} - y_{b,i}|^p]^{\frac{1}{p}},
\end{multline}
where $\lambda$ is a hyperparameter controlling the contribution of $x$ and $y$ towards the cost function, and $p\ge 1$. As $\lambda$ approaches $1$, the optimal transport plan will only map $a$ to $b$ if $a$ are $b$ lie close in the $x$-space. Thus, $p(y|x)$ can be estimated using only the samples drawn from $p(x,y)$. 


\subsection{Further optimizations}
The linear assignment problem is computationally expensive, with the computation cost polynomial in terms of the number of rows (or columns) of the cost matrix. This makes using larger batch-sizes impractical because of associated increased training time. However, larger batch sizes is a requirement when we intend to estimate complicated $\mathbb{P}_R$ distributions (discussed further in Section~\ref{sec:ablation}). Since we only intend to find a transport plan that maps $a \in R$ to $b \in F$ only if $a$ and $b$ lie close in the $x$-space, we can pre-compute the $x$-neighbors for each $a\in R$ (say $k$ nearest neighbors), and only allow a mapping from $a$ to these $x$-neighbors. This corresponds to sparsifying the cost matrix $C$ (or removing high-cost edges from the bipartite graph $R\rightarrow F$). Further, with repeated $(x, y)\in R$ samples drawn from $\mathbb{P}_R$ (in GAN terminology, multiple fake samples are generated per real sample), there are multiple transport plans, with optimal cost. This further allows us to sparsify the cost matrix, by only considering the $k$ \emph{unique} nearest neighbors.  Solving the linear assignment problem with a sparse cost matrix $R\rightarrow F$ can be solved efficiently~\cite{jonker1987shortest}. We call the original approach, as \emph{REgression via Dense Linear Assignment (RE-DLA)}, and the one with sparsified objective as  \emph{REgression via Sparse Linear Assignment (RE-SLA)}. As we see later in Section~\ref{sec:ablation}, having repeated $(x, y)\in R$ samples drawn from $\mathbb{P}_R$ leads to a faster convergence, and RE-SLA provides a computationally inexpensive way to achieve that, which is impractical with RE-DLA. 

One could also try a separate $\lambda$ weight component hyperparameter (automatic relevance determination) for each component of $x$ and $y$. This can potentially lead to even better models. However, we do not explore this option in this paper.

\section{Experimental methodology}\label{sec:eval}
We perform the evaluation in a similar manner as performed in~\cite{aggarwal2019regression}, as described below.
\subsection{Performance assessment metrics}\label{sec:metrics}
Our primary metric of interest is the Negative Log Predictive Density (NLPD). NLPD has an important role in statistical model comparison because of its connection to KL-divergence~\cite{bumham2002model}. Specifically, a lower expected NLPD corresponds to a higher posterior probability. Since our approach models $p(y|x)$ implicitly, it can  only generate the samples of $y$ for each given $x$. Thus, to approximate NLPD, we use Parzen windows\footnote{Whether density needs to be modeled at all depends on the individual application. If only some statistic on $y$ needs to be computed at an $x$, then it can be directly done using samples. If some visualization is needed, drawing some samples would suffice. Only if a closed form representation of $p(y|x)$ is really needed then we have to go for Parzen windows or other density fitting approaches. Also, for NLPD approximation we could explore methods that are less sensitive than Parzen windows.} with Gaussian kernel, which has been used in the GAN literature~\cite{bengio2013better,goodfellow2014generative}. In addition to NLPD, we also report results on the Mean Absolute Error (MAE) and Mean Squared Error (MSE). In terms of sensitivity to outliers, NLPD is most sensitive, followed by MSE, then MAE. Thus, for metric robustness, we ignore the first and fourth quartiles while reporting the metrics. We also experimented with larger \emph{central zones} to compute our metrics, and got the same trend as reported in this paper.

\subsection{Baselines}\label{sec:baselines}
We compare our approach against the following methods.\\
\textbf{CGAN:} Like our approach, CGAN also implicitly models the noise, but is notoriously difficult to train because of the associated minimax optimization. We use the same setup for CGAN as used in~\cite{aggarwal2019regression}.\\
\textbf{Regression with Gaussian Processes (GPs):} GPs have arguably been the workhorse for regression. We provide results for (i) the linear kernel; (ii) the quadratic (Quad) kernel; and (iii) the radial basis function (RBF) kernel.\\
\textbf{Deep Neural Net (DNN):} We use the same DNN implementation as used in~\cite{aggarwal2019regression}.\\
\textbf{eXtreme Gradient BOOSTing (XGBoost):} XGBoost is a decision-tree-based ensemble Machine Learning algorithm that uses a gradient boosting framework. XGBoost is a very popular tool for applied machine learning, and is credited as the driving force for several industry applications. 

\subsection{Datasets}\label{sec:datasets}
We assess the performance of our proposed approaches and compare them with the competing approaches on both synthetic and real-world datasets. To construct the synthetic datasets, we hypothesize how the noise enters the $y$-generation process, and evaluate how well our approaches and the baselines correctly model that noise. We use seven commonly used real-world datasets taken from various repositories. Specifically, we use five datasets from the Delve repository\footnote{https://www.cs.toronto.edu/~delve/data/datasets.html}. Each dataset in the Delve repository has one or more regression tasks associated with it. The datasets and their tasks (if multiple tasks are available) are as follows: Abalone, Bank (32nm task), Census-house (house-price-16H task), Comp-activ (cpu task),  and Pumadyn (32 nm task). We use the Ailerons dataset from the OpenML\footnote{https://www.openml.org/d/296} repository, and CA-housing dataset from scikit-learn\footnote{https://scikit-learn.org/stable/datasets/index.html\#california-housing-dataset}.

\subsection{Parameter selection}
We implement the function $f$, for both RE-DLA and RE-SLA as a feed-forward neural network. We separately feed the input $x$ and noise $z$ through a two-layered network and concatenate the output
representations to pass through another two-layered network. Except for the final layer that employs the linear activation,
we use Rectified linear unit (ReLU) as the activation function, followed by Batch Normalization, which accelerates the training~\cite{ioffe2015batch}. Each hidden layer has 16 neurons. We model $z$ as a one-dimensional Gaussian noise. For the cost-function in (\eqref{eq:cost}), we tried $p\in \{1, 2\}$, and had similar performances. For the experimental results reported in this paper, we use $p=1$. We use ADAM~\cite{kingma2014adam} optimizer for the optimization, with the learning rate set to $0.001$. We use a mini-batch size of $100$ and sample-size of $10$. The sample-size corresponds to number of samples we draw from $\hat{p}(y|x)$, per sample drawn from $p(y|x)$, or number of fake samples generated per real sample. For RE-SLA, we consider $10$ unique nearest neighbors to sparsify the cost matrix. We employ early-stopping to terminate the training of our models if there is no NLPD improvement on a held-out validation set for $10$ epochs. To solve the Linear Assignment Problem (for both RE-DLA and RE-SLA), we use the LAP package\footnote{https://github.com/gatagat/lap}. The hyperparameter $\lambda$ in (\eqref{eq:cost}) is tuned individually for each dataset, based on the NLPD on a held-out validation dataset. Note that, except the hyperparameter $\lambda$, we did not tune any additional hyperparameters. We believe that tuning them would improve the performance of our methods even more; for example, for the synthetic datasets discussed in Section~\ref{sec:synthetic_experiments}, we can achieve a better performance with a less complex architecture. 

For CGAN, we use the same setup as used in~\cite{aggarwal2019regression}. Specifically, we use a six-layer network as the Generator and a three-layer network as the Discriminator. The number of neurons and other parameters are chosen individually for different datasets. For DNN, we use the same seven-layer network as used in~\cite{aggarwal2019regression}, which has a comparable architecture to CGAN. For GP, we use the GPy package~\cite{gpy2014} with automatic hyperparameter optimization. For XGboost we use the XGboost package~\cite{xgboost}. The GP, DNN, and XGBoost assume additive Gaussian noise model (we use MSE loss for DNN and XGBoost). While GPs directly estimate $p(y|x)$, for DNN and XGboost we assume a Gaussian likelihood and choose the variance to optimize NLPD on a validation set in order to calculate $p(y|x)$. For CGAN, RE-DLA and RE-SLA, we generate $2,000$ samples to estimate NLPD using Parzen windows, as mentioned earlier in Section~\ref{sec:metrics}. A large number of samples leads to a stable NLPD computation. The experiments reported in this paper are run on Intel i5-8400 CPU with 6 cores, without GPU support.

\subsection{Methodology}\label{sec:datasets}
We report the results using five-fold cross validation, and also report the associated standard deviations. For all experiments, both the input features $x$ and output $y$ were scaled to have overall zero mean and unit variance. The results (NLPD, MAE and MSE) are reported on the scaled output $y$. The statistical significance tests were conducted using the method proposed by~\cite{demvsar2006statistical}, specifically using the Friedman test with Nemenyi post-hoc analysis. 

\section{Results and discussion}
\subsection{Synthetic experiments}\label{sec:synthetic_experiments}
We constructed synthetic datasets, by hypothesizing how the noise enters the $y$-generation process, and evaluate how well our approaches and the baselines correctly model that noise. Note that, in methods that explicitly model a specific noise form, a different explicit model is required for a different noise form. Such explicit models are expected to perform better on the noise forms for which they are designed. But the important point is that implicit models such as RE-DLA and RE-SLA are able to use a generic architecture to model each of the noise forms accurately, as we will see in the next sections. \\

\begin{figure}[!t]
\centering     
\subfigure{\label{fig:sinus:real}\includegraphics[width=0.32\linewidth]{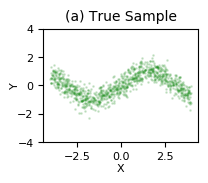}} 
\\
\subfigure{\label{fig:sinus:cgan}\includegraphics[width=0.32\linewidth]{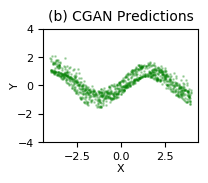}}\hfil 
\subfigure{\label{fig:sinus:redla}\includegraphics[width=0.32\linewidth]{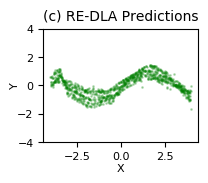}}\hfil 
\subfigure{\label{fig:sinus:resla}\includegraphics[width=0.32\linewidth]{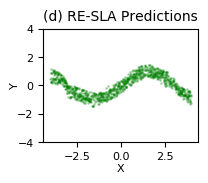}}\hfil 
\subfigure{\label{fig:sinus:gp}\includegraphics[width=0.32\linewidth]{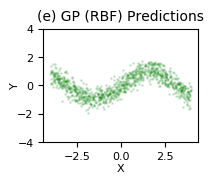}}\hfil 
\subfigure{\label{fig:sinus:xgb}\includegraphics[width=0.32\linewidth]{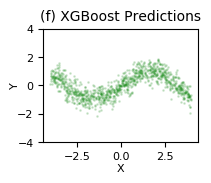}}\hfil 
\subfigure{\label{fig:sinus:dnn}\includegraphics[width=0.31\linewidth]{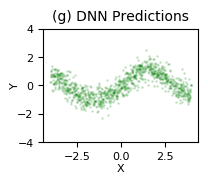}}\hfil 
\caption{\texttt{sinus} dataset: Generated samples.}
\label{fig:sinus}
\end{figure}

\textbf{Sinusodial map with additive gaussian noise} (\texttt{sinus}): We first generate a dataset with a standard Gaussian noise but $y$ as a non-linear function of $x$ as $y = sin(x) + z$, where $z \sim \mathcal{N}(0,1), x \sim \mathcal{U}[-4,4]$ and $\mathcal{U}$ is the uniform distribution. Figure~\ref{fig:sinus} shows the generated samples of the true model, and the predicted sample clouds of various regression approaches. Clearly, the sample clouds produced by GP (except linear kernel, which by design, cannot identify the non-linear relation), XGBoost and DNN look more similar to the true samples as compared to the implicit noise-modeling approaches like CGAN, RE-DLA and RE-SLA. The reason is because this dataset precisely corresponds to the assumptions of the GP, XGBoost and DNN. Table~\ref{tab:sinus_results} lists NLPD, MAE and MSE metric values for all the methods on the test data. There is not a significant difference between the methods on various metrics.\\

\begin{table}[t]
\caption{Performance comparison on the \texttt{sinus} dataset.}
\label{tab:sinus_results}
\begin{center}
\begin{scriptsize}
\begin{sc}
\begin{threeparttable}
\begin{tabularx}{\linewidth}{lRRR}
\toprule
Method & NLPD & MAE & MSE\\
\midrule
RE-DLA &  $ 0.390 \pm 0.029 $  &  $ 0.364 \pm 0.004 $  &  $ 0.147 \pm 0.005 $ \\
RE-SLA &  $ 0.374 \pm 0.025 $  &  $ 0.366 \pm 0.004 $  &  $ 0.147 \pm 0.005 $ \\
CGAN &  $ 0.646 \pm 0.062 $  &  $ 0.406 \pm 0.025 $  &  $ 0.180 \pm 0.017 $ \\
GPR (RBF) &  $ 0.539 \pm 0.011 $  &  $ 0.361 \pm 0.006 $  &  $ 0.146 \pm 0.005 $ \\
GPR (Quad) &  $ 0.539 \pm 0.011 $  &  $ 0.361 \pm 0.006 $  &  $ 0.146 \pm 0.005 $ \\
GPR (Linear) &  $ 1.220 \pm 0.016 $  &  $ 0.761 \pm 0.019 $  &  $ 0.628 \pm 0.030 $ \\
DNN &  $ 0.545 \pm 0.015 $  &  $ 0.363 \pm 0.006 $  &  $ 0.147 \pm 0.005 $ \\
XGBoost &  $ 0.548 \pm 0.017 $  &  $ 0.365 \pm 0.006 $  &  $ 0.149 \pm 0.005 $ \\
\bottomrule
\end{tabularx}
\end{threeparttable}
\end{sc}
\end{scriptsize}
\end{center}
\end{table}

\begin{figure}[!t]
\centering     
\subfigure{\label{fig:exp:real}\includegraphics[width=0.32\linewidth]{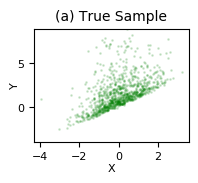}} 
\\
\subfigure{\label{fig:exp:cgan}\includegraphics[width=0.32\linewidth]{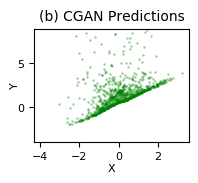}}\hfil 
\subfigure{\label{fig:exp:redla}\includegraphics[width=0.32\linewidth]{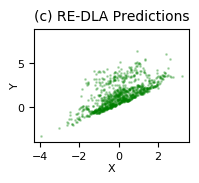}}\hfil 
\subfigure{\label{fig:exp:resla}\includegraphics[width=0.32\linewidth]{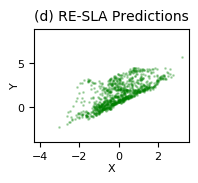}}\hfil 
\subfigure{\label{fig:exp:gp}\includegraphics[width=0.32\linewidth]{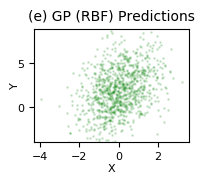}}\hfil 
\subfigure{\label{fig:exp:xgb}\includegraphics[width=0.32\linewidth]{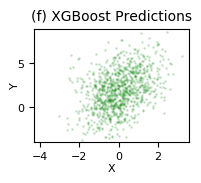}}\hfil 
\subfigure{\label{fig:exp:dnn}\includegraphics[width=0.31\linewidth]{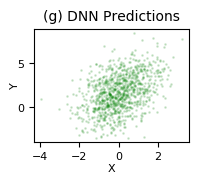}}\hfil 
\caption{\texttt{exp} dataset: Generated samples.}
\label{fig:exp}
\end{figure}

\textbf{Linear map with additive exponential noise }(\texttt{exp}):  Next, we consider a more complex generation process, which involves an additive exponential noise. We employ the following generation process: $y = x + \exp(z)$ where $x, z \sim \mathcal{N}(0,1)$. From Figure~\ref{fig:exp}, we observe that CGAN, RE-DLA and RE-SLA easily adopt the asymmetry while the other approaches are unsuccessful in doing so. Further, as shown in Table~\ref{tab:exp_results}, RE-DLA and RE-SLA outperform CGAN by a large margin.\\

\begin{figure}[!t]
\centering     
\subfigure{\label{fig:hetero:real}\includegraphics[width=0.32\linewidth]{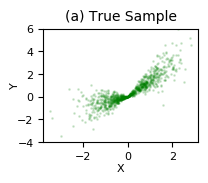}} 
\\
\subfigure{\label{fig:hetero:cgan}\includegraphics[width=0.32\linewidth]{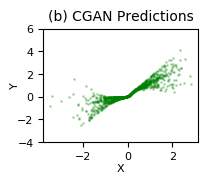}}\hfil 
\subfigure{\label{fig:hetero:redla}\includegraphics[width=0.32\linewidth]{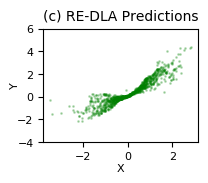}}\hfil 
\subfigure{\label{fig:hetero:resla}\includegraphics[width=0.32\linewidth]{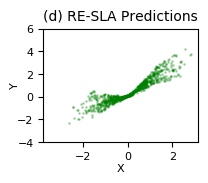}}\hfil 
\subfigure{\label{fig:hetero:gp}\includegraphics[width=0.32\linewidth]{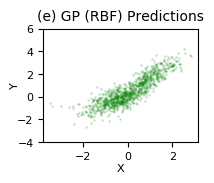}}\hfil 
\subfigure{\label{fig:hetero:xgb}\includegraphics[width=0.32\linewidth]{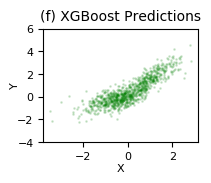}}\hfil 
\subfigure{\label{fig:hetero:dnn}\includegraphics[width=0.31\linewidth]{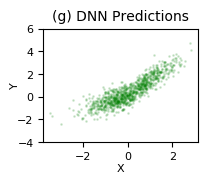}}\hfil 
\caption{\texttt{heteroscedastic} dataset: : Generated samples.}
\label{fig:hetero}
\end{figure}

\begin{table}[t]
\caption{Performance comparison on the \texttt{exp} dataset.}
\label{tab:exp_results}
\begin{center}
\begin{scriptsize}
\begin{sc}
\begin{threeparttable}
\begin{tabularx}{\linewidth}{lRRR}
\toprule
Method & NLPD & MAE & MSE\\
\midrule
RE-DLA &  $ 0.138 \pm 0.082 $  &  $ 0.232 \pm 0.021 $  &  $ 0.104 \pm 0.029 $ \\
RE-SLA &  $ 0.084 \pm 0.075 $  &  $ 0.235 \pm 0.016 $  &  $ 0.075 \pm 0.012 $ \\
CGAN &  $ 0.328 \pm 0.106 $  &  $ 0.258 \pm 0.027 $  &  $ 0.160 \pm 0.070 $ \\
GPR (RBF) &  $ 0.935 \pm 0.005 $  &  $ 0.400 \pm 0.028 $  &  $ 0.169 \pm 0.024 $ \\
GPR (Quad) &  $ 0.935 \pm 0.005 $  &  $ 0.400 \pm 0.028 $  &  $ 0.169 \pm 0.024 $ \\
GPR (Linear) &  $ 0.935 \pm 0.006 $  &  $ 0.400 \pm 0.026 $  &  $ 0.169 \pm 0.023 $ \\
DNN &  $ 0.916 \pm 0.213 $  &  $ 0.402 \pm 0.026 $  &  $ 0.170 \pm 0.022 $ \\
XGBoost &  $ 0.949 \pm 0.197 $  &  $ 0.439 \pm 0.025 $  &  $ 0.203 \pm 0.023 $ \\
\bottomrule
\end{tabularx}
\end{threeparttable}
\end{sc}
\end{scriptsize}
\end{center}
\end{table}
\textbf{Linear map with heteroscedastic noise} (\texttt{heteroscedastic}): We generate a dataset with a heteroschedastic noise, i.e., the noise that is dependent on $x$. Specifically, we generate the dataset as following: $y = x + h(x, z)$ where $h(x, z) = (0.001 + 0.5 |x|) \times z$ and $z \sim \mathcal{N}(1,1)$. Around $x=0$, the noise is small, and the noise keeps increasing as we go away from $x=0$. Figure~\ref{fig:hetero} shows the true sample distribution and the samples generated by various methods. The implicit noise-modeling approaches, i.e., CGAN, RE-DLA and RE-SLA generate much better sample clouds as compared to the other approaches.
Furthermore, as shown in Table~\ref{tab:hetero_results}, CGAN, RE-DLA and RE-SLA achieve much better NLPDs as compared to the other approaches. Among the implicit methods, RE-DLA and RE-SLA outperform the CGAN. 

\begin{table}[t]
\caption{Performance on the \texttt{heteroscedastic} dataset.}
\label{tab:hetero_results}
\begin{center}
\begin{scriptsize}
\begin{sc}
\begin{threeparttable}
\begin{tabularx}{\linewidth}{Xrrr}
\toprule
Method & NLPD & MAE & MSE\\
\midrule
RE-DLA &  $ -0.190 \pm 0.073 $  &  $ 0.179 \pm 0.009 $  &  $ 0.040 \pm 0.005 $ \\
RE-SLA &  $ -0.186 \pm 0.071 $  &  $ 0.178 \pm 0.010 $  &  $ 0.040 \pm 0.005 $ \\
CGAN &  $ 0.122 \pm 0.084 $  &  $ 0.196 \pm 0.012 $  &  $ 0.044 \pm 0.005 $ \\
GPR (RBF) &  $ 0.183 \pm 0.014 $  &  $ 0.177 \pm 0.011 $  &  $ 0.040 \pm 0.005 $ \\
GPR (Quad) &  $ 0.183 \pm 0.014 $  &  $ 0.177 \pm 0.011 $  &  $ 0.040 \pm 0.005 $ \\
GPR (Lin) &  $ 0.397 \pm 0.011 $  &  $ 0.280 \pm 0.007 $  &  $ 0.083 \pm 0.004 $ \\
DNN &  $ 0.191 \pm 0.058 $  &  $ 0.180 \pm 0.010 $  &  $ 0.041 \pm 0.005 $ \\
XGBoost &  $ 0.206 \pm 0.055 $  &  $ 0.180 \pm 0.011 $  &  $ 0.041 \pm 0.005 $ \\
\bottomrule
\end{tabularx}
\end{threeparttable}
\end{sc}
\end{scriptsize}
\end{center}
\end{table}

\textbf{Multi-modal map and noise} (\texttt{multi-modal}): Next, we generate a more complex dataset with a multi-modal $p(y|x)$ distribution which also changes with $x$. Such distributions can occur in real world phenomena such as dynamical systems which switch between multiple states depending on latent factors like temperature~\cite{einbeck2006modelling}; this can create scenarios where the same input $x$ can be mapped to two values of $y$. 
We use the following procedure to generate a multi-modal data where $y$ is: $1.2 x + 0.03 z$ or $x + 0.6 + 0.03 z$ when $0.4 < x$; $0.5 x + 0.01 z$ or $0.6 x + 0.01 z$ when $0.4 \leq x < 0.6$ and; $0.5 + 0.02 z$ when $0.6\leq x$, with $z \sim \mathcal{N}(0,1)$ and $x \sim \mathcal{U}[0,1]$. Figure~\ref{fig:multi} shows the true sample distribution and the sample clouds estimated by various approaches. As expected, CGAN, RE-DLA and RE-SLA are able to model the complex multi-modal distribution, while the other approaches only estimate a Unimodal Gaussian distribution for a given $x$. 
Finally, as shown in Table~\ref{tab:multi_results}, RE-DLA and RE-SLA outperform CGAN. 

\begin{figure}[!t]
\centering     
\subfigure{\label{fig:multi:real}\includegraphics[width=0.32\linewidth]{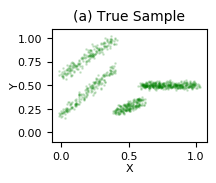}} 
\\
\subfigure{\label{fig:multi:cgan}\includegraphics[width=0.32\linewidth]{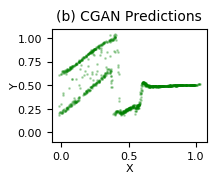}}\hfil 
\subfigure{\label{fig:multi:redla}\includegraphics[width=0.32\linewidth]{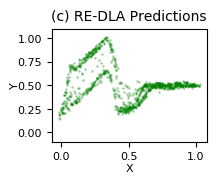}}\hfil 
\subfigure{\label{fig:multi:resla}\includegraphics[width=0.32\linewidth]{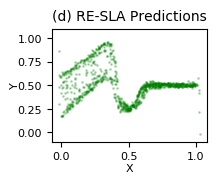}}\hfil 
\subfigure{\label{fig:multi:gp}\includegraphics[width=0.32\linewidth]{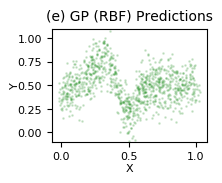}}\hfil 
\subfigure{\label{fig:multi:xgb}\includegraphics[width=0.32\linewidth]{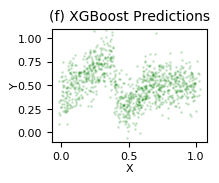}}\hfil 
\subfigure{\label{fig:multi:dnn}\includegraphics[width=0.32\linewidth]{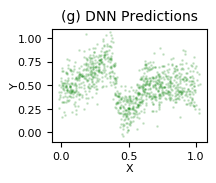}}\hfil 
\caption{\texttt{multi-modal} dataset: Generated samples.}
\label{fig:multi}
\end{figure}

\begin{table}[t]
\caption{Performance on the \texttt{multi-modal} dataset.}
\label{tab:multi_results}
\begin{center}
\begin{scriptsize}
\begin{sc}
\begin{threeparttable}
\begin{tabularx}{\linewidth}{Xrrr}
\toprule
Method & NLPD & MAE & MSE\\
\midrule
RE-DLA &  $ -0.241 \pm 0.030 $  &  $ 0.313 \pm 0.035 $  &  $ 0.219 \pm 0.016 $ \\
RE-SLA &  $ -0.195 \pm 0.138 $  &  $ 0.337 \pm 0.024 $  &  $ 0.213 \pm 0.031 $ \\
CGAN &  $ -0.172 \pm 0.099 $  &  $ 0.269 \pm 0.041 $  &  $ 0.206 \pm 0.017 $ \\
GPR (RBF) &  $ 0.825 \pm 0.010 $  &  $ 0.391 \pm 0.015 $  &  $ 0.239 \pm 0.013 $ \\
GPR (Quad) &  $ 0.824 \pm 0.010 $  &  $ 0.390 \pm 0.016 $  &  $ 0.238 \pm 0.014 $ \\
GPR (Lin) &  $ 1.163 \pm 0.025 $  &  $ 0.609 \pm 0.031 $  &  $ 0.500 \pm 0.048 $ \\
DNN &  $ 0.817 \pm 0.018 $  &  $ 0.385 \pm 0.016 $  &  $ 0.231 \pm 0.014 $ \\
XGBoost &  $ 0.815 \pm 0.017 $  &  $ 0.381 \pm 0.010 $  &  $ 0.227 \pm 0.011 $ \\
\bottomrule
\end{tabularx}
\end{threeparttable}
\end{sc}
\end{scriptsize}
\end{center}
\end{table}

\begin{table*}[!t]
\caption{NLPD for various methods and datasets.}
\label{tab:real_nlpd}
\begin{center}
\begin{scriptsize}
\begin{sc}
\begin{threeparttable}
\begin{tabularx}{\linewidth}{Xrrrrrrr}
\toprule
Method & Abalone & Ailerons & Bank & CA-housing & Census-house & Comp-activ & Pumadyn\\
\midrule
RE-DLAP &  $ 0.403 \pm 0.057 $  &  $ 0.014 \pm 0.013 $  &  $ -0.021 \pm 0.020 $  &  $ 0.314 \pm 0.180 $  &  $ -0.408 \pm 0.022 $  &  $ -0.983 \pm 0.032 $  &  $ -0.128 \pm 0.080 $ \\
RE-SLAP &  $ 0.420 \pm 0.049 $  &  $ 0.013 \pm 0.013 $  &  $ 0.004 \pm 0.042 $  &  $ 0.320 \pm 0.169 $  &  $ -0.431 \pm 0.010 $  &  $ -0.987 \pm 0.033 $  &  $ -0.113 \pm 0.085 $ \\
CGAN &  $ 1.030 \pm 0.310 $  &  $ 0.112 \pm 0.034 $  &  $ 0.167 \pm 0.024 $  &  $ 0.349 \pm 0.167 $  &  $ -0.343 \pm 0.037 $  &  $ -0.942 \pm 0.034 $  &  $ 0.348 \pm 0.064 $ \\
GPR (RBF) &  $ 0.656 \pm 0.014 $  &  $ 0.155 \pm 0.007 $  &  $ 0.235 \pm 0.004 $  &  $ 0.476 \pm 0.070 $  &  $ 0.489 \pm 0.009 $  &  $ -0.872 \pm 0.010 $  &  $ 0.988 \pm 0.013 $ \\
GPR (Quad) &  $ 0.654 \pm 0.015 $  &  $ 0.149 \pm 0.005 $  &  $ 0.229 \pm 0.005 $  &  $ 0.468 \pm 0.056 $  &  $ 0.490 \pm 0.009 $  &  $ -0.882 \pm 0.006 $  &  $ 0.988 \pm 0.013 $ \\
GPR (Lin) &  $ 0.722 \pm 0.021 $  &  $ 0.264 \pm 0.007 $  &  $ 0.408 \pm 0.005 $  &  $ 0.664 \pm 0.039 $  &  $ 0.839 \pm 0.004 $  &  $ 0.394 \pm 0.032 $  &  $ 1.048 \pm 0.008 $ \\
DNN &  $ 0.678 \pm 0.049 $  &  $ 0.210 \pm 0.035 $  &  $ 0.240 \pm 0.023 $  &  $ 0.580 \pm 0.067 $  &  $ 0.582 \pm 0.025 $  &  $ -0.665 \pm 0.044 $  &  $ 0.855 \pm 0.166 $ \\
XGBoost &  $ 0.698 \pm 0.027 $  &  $ 0.171 \pm 0.019 $  &  $ 0.270 \pm 0.025 $  &  $ 0.573 \pm 0.025 $  &  $ 0.508 \pm 0.022 $  &  $ -0.972 \pm 0.043 $  &  $ -0.044 \pm 0.027 $ \\
\bottomrule
\end{tabularx}
\begin{tablenotes}
 \item[*] The performance improvement of RE-DLA and RE-SLA is statistically significant with respect to the other methods. 
\end{tablenotes}
\end{threeparttable}
\end{sc}
\end{scriptsize}
\end{center}
\end{table*}

\begin{table*}[!t]
\caption{MAE for various methods and datasets.}
\label{tab:real_mae}
\begin{center}
\begin{scriptsize}
\begin{sc}
\begin{threeparttable}
\begin{tabularx}{\linewidth}{Xrrrrrrr}
\toprule
Method & Abalone & Ailerons & Bank & CA-housing & Census-house & Comp-activ & Pumadyn\\
\midrule
RE-DLAP &  $ 0.340 \pm 0.021 $  &  $ 0.229 \pm 0.003 $  &  $ 0.211 \pm 0.005 $  &  $ 0.300 \pm 0.043 $  &  $ 0.146 \pm 0.003 $  &  $ 0.079 \pm 0.004 $  &  $ 0.201 \pm 0.015 $ \\
RE-SLAP &  $ 0.346 \pm 0.017 $  &  $ 0.229 \pm 0.003 $  &  $ 0.218 \pm 0.011 $  &  $ 0.308 \pm 0.047 $  &  $ 0.142 \pm 0.002 $  &  $ 0.079 \pm 0.004 $  &  $ 0.204 \pm 0.018 $ \\
CGAN &  $ 0.563 \pm 0.156 $  &  $ 0.226 \pm 0.004 $  &  $ 0.255 \pm 0.005 $  &  $ 0.299 \pm 0.056 $  &  $ 0.150 \pm 0.007 $  &  $ 0.084 \pm 0.004 $  &  $ 0.324 \pm 0.021 $ \\
GPR (RBF) &  $ 0.357 \pm 0.012 $  &  $ 0.225 \pm 0.004 $  &  $ 0.248 \pm 0.003 $  &  $ 0.330 \pm 0.030 $  &  $ 0.187 \pm 0.003 $  &  $ 0.081 \pm 0.002 $  &  $ 0.557 \pm 0.015 $ \\
GPR (Quad) &  $ 0.357 \pm 0.013 $  &  $ 0.223 \pm 0.003 $  &  $ 0.244 \pm 0.004 $  &  $ 0.317 \pm 0.029 $  &  $ 0.177 \pm 0.003 $  &  $ 0.078 \pm 0.001 $  &  $ 0.557 \pm 0.015 $ \\
GPR (Lin) &  $ 0.388 \pm 0.023 $  &  $ 0.252 \pm 0.004 $  &  $ 0.302 \pm 0.005 $  &  $ 0.392 \pm 0.025 $  &  $ 0.302 \pm 0.005 $  &  $ 0.248 \pm 0.016 $  &  $ 0.586 \pm 0.009 $ \\
DNN &  $ 0.355 \pm 0.018 $  &  $ 0.234 \pm 0.004 $  &  $ 0.236 \pm 0.004 $  &  $ 0.333 \pm 0.022 $  &  $ 0.190 \pm 0.004 $  &  $ 0.092 \pm 0.002 $  &  $ 0.487 \pm 0.075 $ \\
XGBoost &  $ 0.367 \pm 0.019 $  &  $ 0.222 \pm 0.002 $  &  $ 0.247 \pm 0.007 $  &  $ 0.347 \pm 0.010 $  &  $ 0.184 \pm 0.005 $  &  $ 0.068 \pm 0.001 $  &  $ 0.203 \pm 0.006 $ \\
\bottomrule
\end{tabularx}
\begin{tablenotes}
 \item[*] The performance improvement of RE-DLA and RE-SLA is statistically significant with respect to the other methods. 
\end{tablenotes}
\end{threeparttable}
\end{sc}
\end{scriptsize}
\end{center}
\end{table*}

\begin{table*}[!t]
\caption{MSE for various methods and datasets.}
\label{tab:real_mse}
\begin{center}
\begin{scriptsize}
\begin{sc}
\begin{threeparttable}
\begin{tabularx}{\linewidth}{Xrrrrrrr}
\toprule
Method & Abalone & Ailerons & Bank & CA-housing & Census-house & Comp-activ & Pumadyn\\
\midrule
RE-DLAP &  $ 0.131 \pm 0.015 $  &  $ 0.060 \pm 0.002 $  &  $ 0.055 \pm 0.002 $  &  $ 0.109 \pm 0.034 $  &  $ 0.026 \pm 0.001 $  &  $ 0.007 \pm 0.001 $  &  $ 0.045 \pm 0.007 $ \\
RE-SLAP &  $ 0.135 \pm 0.013 $  &  $ 0.059 \pm 0.002 $  &  $ 0.058 \pm 0.005 $  &  $ 0.110 \pm 0.032 $  &  $ 0.025 \pm 0.001 $  &  $ 0.007 \pm 0.001 $  &  $ 0.046 \pm 0.008 $ \\
CGAN &  $ 0.459 \pm 0.238 $  &  $ 0.058 \pm 0.003 $  &  $ 0.081 \pm 0.004 $  &  $ 0.113 \pm 0.042 $  &  $ 0.028 \pm 0.003 $  &  $ 0.008 \pm 0.001 $  &  $ 0.117 \pm 0.016 $ \\
GPR (RBF) &  $ 0.144 \pm 0.010 $  &  $ 0.059 \pm 0.002 $  &  $ 0.069 \pm 0.002 $  &  $ 0.126 \pm 0.023 $  &  $ 0.042 \pm 0.001 $  &  $ 0.007 \pm 0.000 $  &  $ 0.351 \pm 0.017 $ \\
GPR (Quad) &  $ 0.144 \pm 0.012 $  &  $ 0.057 \pm 0.001 $  &  $ 0.067 \pm 0.002 $  &  $ 0.117 \pm 0.021 $  &  $ 0.038 \pm 0.001 $  &  $ 0.007 \pm 0.000 $  &  $ 0.351 \pm 0.017 $ \\
GPR (Lin) &  $ 0.168 \pm 0.018 $  &  $ 0.072 \pm 0.002 $  &  $ 0.102 \pm 0.003 $  &  $ 0.171 \pm 0.019 $  &  $ 0.104 \pm 0.003 $  &  $ 0.070 \pm 0.008 $  &  $ 0.393 \pm 0.011 $ \\
DNN &  $ 0.143 \pm 0.014 $  &  $ 0.062 \pm 0.002 $  &  $ 0.063 \pm 0.002 $  &  $ 0.125 \pm 0.016 $  &  $ 0.041 \pm 0.002 $  &  $ 0.009 \pm 0.000 $  &  $ 0.275 \pm 0.074 $ \\
XGBoost &  $ 0.151 \pm 0.014 $  &  $ 0.056 \pm 0.001 $  &  $ 0.070 \pm 0.004 $  &  $ 0.135 \pm 0.008 $  &  $ 0.040 \pm 0.002 $  &  $ 0.005 \pm 0.000 $  &  $ 0.046 \pm 0.003 $ \\
\bottomrule
\end{tabularx}
\begin{tablenotes}
 \item[*] The performance improvement of RE-DLA and RE-SLA is statistically significant with respect to the other methods. 
\end{tablenotes}
\end{threeparttable}
\end{sc}
\end{scriptsize}
\end{center}
\end{table*}

\subsection{Real-world datasets}
Table~\ref{tab:real_nlpd} shows the performance statistics of various methods on various real-world datasets on the NLPD metric, which is our primary metric of interest. We observe that the implicit noise-modeling approaches (RE-DLA, RE-SLA and CGAN) generally outperform the other methods. Morover, RE-DLA and RE-SLA consistently performing better than all the competing approaches on all the datasets by a large margin. As in the synthetic experiments, this improvement is even more appreciable because of the fact that the RE-DLA and RE-SLA employ the same generic architecture for all experiments, without any hyperparameter optimization as compared to CGAN which selects its hyperparameters independently for each experiment. The performance improvement of RE-DLA and RE-SLA is statistically significant with respect to the other methods, as explained later. We explore RE-DLA and RE-SLA further in Section~\ref{sec:ablation}.

Table~\ref{tab:real_mae} and Table~\ref{tab:real_mse} show the performance statistics of various methods on the various real-world datasets on the MAE and MSE metrics, respectively. On four out of seven datasets (Abalone, Bank, Census-house and Pumadyn), the proposed approaches RE-DLA and RE-SLA outperform the competing approaches on both MAE and MSE metrics. Even on the other datasets, RE-DLA and RE-SLA perform on par with the state-of-the-art methods.  The performance improvement of RE-DLA and RE-SLA is statistically significant.
\\\textbf{Statistical significance test:} We analysed the statistical significance of the results of various methods using the Friedman test with Nemenyi post-hoc analysis. 
When comparing eight methods (precisely our case), for a confidence level of $p = 0.05$, method $a$'s performance is statistically significant than method $b$'s performance if the corresponding average ranks differ by at least $10.5/\sqrt{N}$, where $N$ is the number of domains on which the comparison is made. As such, the value of $N$ should be large enough to have convincing statistical significance evaluation. Thus, we treat each test instance (and not dataset) as a domain, and test the statistical significance with respect to that, the corresponding metrics being individual NLPD, absolute error and squared error. For all the three metrics, Friedman test says that the methods are not equivalent with small $p$-value. According to the Nemenyi test, the difference between the RE-DLA and RE-SLA performances is not statistically significant, while the performance of both RE-DLA and RE-SLA is statistically significant over other methods  with small $p$-value.

\subsection{RE-DLA vs RE-SLA}\label{sec:ablation}
In this section, we conduct experiments to understand the advantages of RE-SLA over RE-DLA. As mentioned earlier, RE-DLA is prohibitive on large mini-batch sizes. However, estimating complicated distributions requires large number of samples, and hence larger mini-batch sizes. RE-SLA is a computationally inexpensive solution that approximates the optimal transport cost by sparsifying the cost matrix, i.e., by only considering the $k$-nearest unique neighbors for each $a\in R$ and $b\in F$. Here, the word, \emph{unique} refers to the case in which we draw multiple samples from the estimated distribution $\hat{p}(y|x)$, for each unique sample drawn from the real distribution $p(y|x)$. The sample-size in this analysis refers to the number of samples drawn from the distribution $\hat{p}(y|x)$, for each unique sample drawn from the real distribution $p(y|x)$. To illustrate the need for larger mini-batch and sample sizes, we synthetically construct a dataset which admits a complicated distribution. Specifically, we generate $5,000$ data-points, with two-dimensional $x$ and one dimensional $y$. The $x$ are randomly sampled from $200$ random Gaussian distributions. At a given $x$, $y$ is the value of the Gaussian mixture probability density, with a small Gaussian noise added. 

\begin{figure}[!t]
\vskip 0.2in
\centering     
\subfigure[]{\label{fig:ablation:batches}\includegraphics[width=0.50\linewidth]{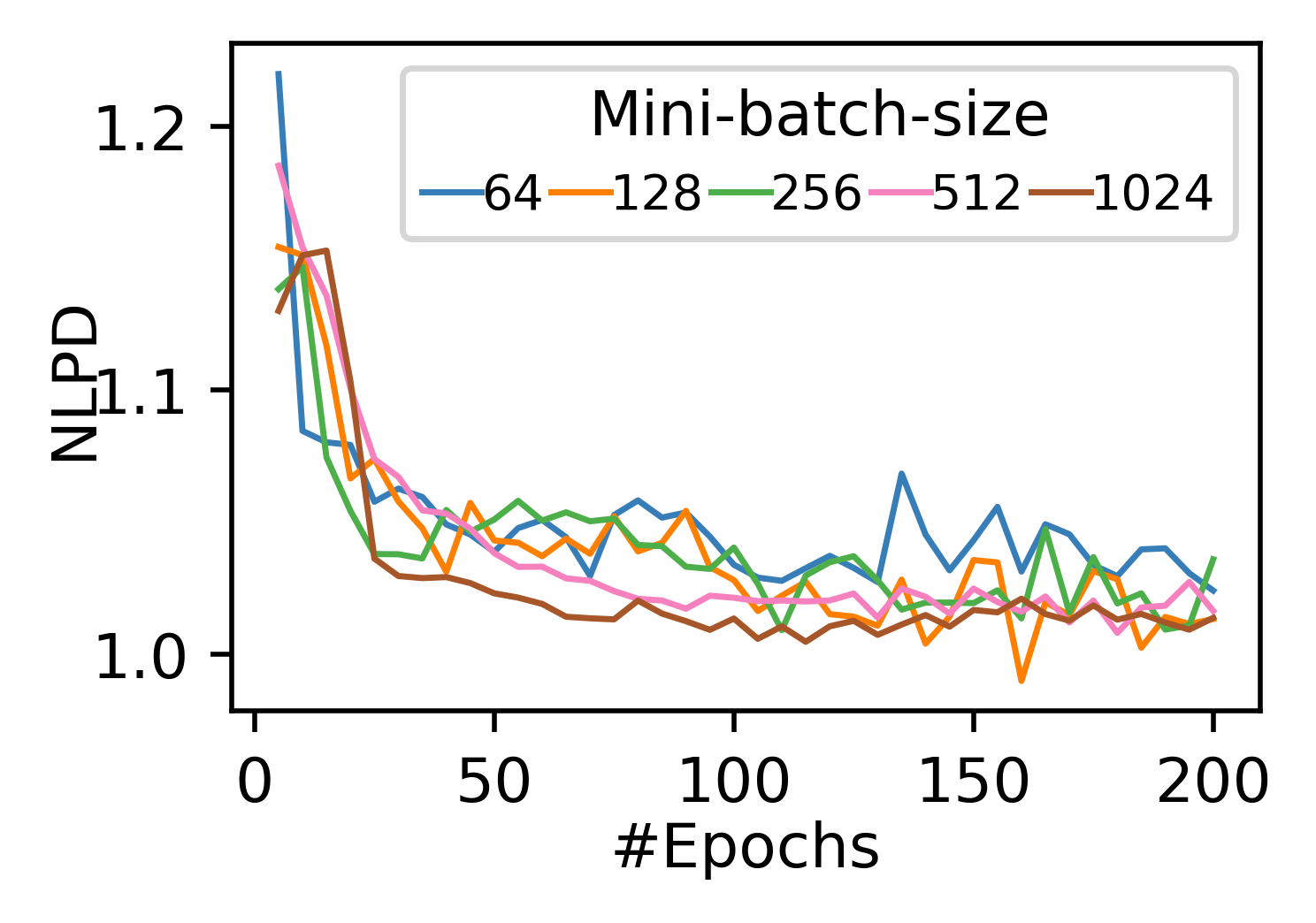}}\hfil 
\subfigure[]{\label{fig:ablation:samples}\includegraphics[width=0.50\linewidth]{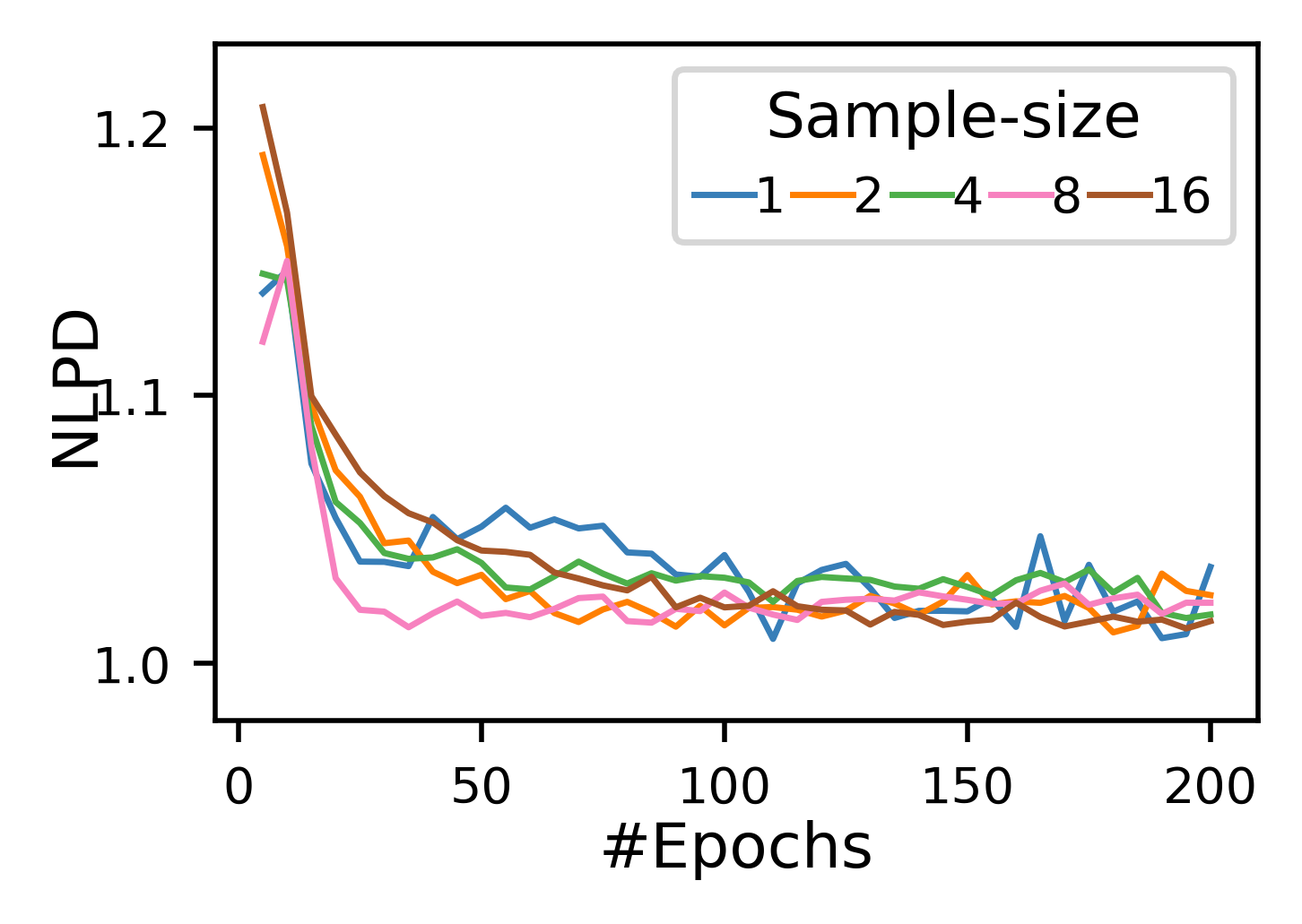}}\hfil 
\caption{Variation of the NLPD with mini-batch size and sample-size. For the mini-batch size experiment, the sample-size is kept fixed at $1$, while for the sample-size experiment, the mini-batch size is kept fixed at $256$.}
\label{fig:multi}
\end{figure}

In the first experiment, we investigate how the NLPD varies with the mini-batch size; and in the second experiment, we investigate how the NLPD varies with the sample size. The results for these experiments is shown in Figures~\ref{fig:ablation:batches} and Figures~\ref{fig:ablation:samples}, respectively. For both experiments, we see that, as the mini-batch size and sample-size are increased, the NLPD values tend to get better and become stable. Thus, using larger mini-batches and multiple samples is of advantage, when modeling complicated distributions. 

\begin{figure}[!t]
\begin{center}
\centerline{\includegraphics[width=0.9\columnwidth]{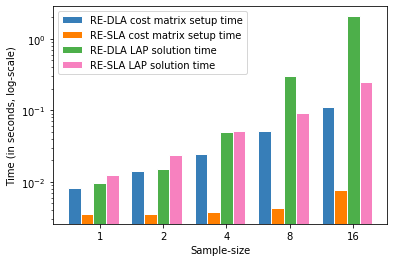}}
\caption{Average runtime per iteration of RE-DLA and RE-SLA for different sample sizes (mini-batch size = $256$).}
\label{fig:ablation:runtime}
\end{center}
\vskip -0.2in
\end{figure}

To show the advantage of RE-SLA over RE-DLA, we plot the average runtime per iteration of RE-DLA and RE-SLA for different sample-sizes. We decompose the runtime into two parts: the cost matrix setup time, which corresponds to computing the entries of the cost-matrix $C$; and the Linear Assignment Problem (LAP) solution time. 
As shown in Figure~\ref{fig:ablation:runtime}, RE-SLA is able to use even larger sample-sizes at a fraction of the time of RE-DLA. For example, when using a sample-size of $16$, the LAP solution time for RE-SLA is $~10\times$ less than that of RE-DLA. The run-times with different batch-sizes also show a similar trend. RE-SLA also has less cost-matrix setup cost, which is expected as RE-SLA has to compute much smaller number of entries in the cost-matrix as compared to RE-DLA. However, setup-time is not a matter of concern, as the dense cost-matrix can be expressed as matrix operations on the input features matrices, which can leverage GPUs.


\section{Conclusion}
The main takeaways from this paper are: (a) we proposed an implicit model for regression which directly optimizes the optimal transport cost between the true probability distribution and the estimated distribution and does not suffer from the issues associated with the minimax approach; (b) our approach, though using a basic setup without any extensive hyperparameter optimization outperforms the competing approaches; (c) we propose a computationally efficient variant of our approach that scales to larger min-batch sizes. Though we only experiment with one-dimensional $y$ in this paper, our proposed approaches are easily applicable to multi-dimensional $y$. In addition, there are other avenues of optimization of our approaches worth investigating, for example, other ways to approximate the optimal transport cost, such as the greedy approach; and automatic relevance determination via choosing different scaling weights for each component of $x$ and $y$. It is important to perform further benchmarking on applications involving complex noise forms. This constitutes an important future work.

\bibliography{example_paper}
\bibliographystyle{icml2020}





\end{document}